%% file: main.tex
\documentclass[11pt,a4paper]{article}

\usepackage[margin=1in]{geometry}
\usepackage[T1]{fontenc}
\usepackage[utf8]{inputenc}
\usepackage{lmodern}
\usepackage{microtype}

\usepackage{amsmath,amssymb,amsfonts,bm}

\usepackage{graphicx}
\graphicspath{{figures/}}
\usepackage{float}
\usepackage{caption}
\usepackage{subcaption}
\usepackage{multirow,makecell,tabularx,adjustbox,booktabs}
\usepackage{cellspace}
\usepackage{comment}
\usepackage[dvipsnames]{xcolor}

\newcommand\todo[1]{#1}

\usepackage[numbers,sort&compress]{natbib}
\usepackage[colorlinks=true,linkcolor=MidnightBlue,citecolor=MidnightBlue,urlcolor=MidnightBlue]{hyperref}

\title{\textbf{Probabilistic Ship Trajectory Forecasting using\\Transformer Models on AIS Data}}
\author{Frederik E.\ T.\ Sch\"oller\thanks{This work was carried out while the author was with the Automation and Control Group, Department of Electrical Engineering, Technical University of Denmark, Kgs.\ Lyngby, Denmark. The author is now an independent researcher.}\\[4pt]
  \small Technical University of Denmark\\
  \small \texttt{frederikschoeller@gmail.com}}
\date{\small Manuscript written in 2021; posted as a preprint in 2026.}

\begin{document}
\maketitle

\begin{abstract}
\noindent Anticipation of vessel trajectories is an essential task when navigating at sea. This applies to both human navigators and autonomous surface vessels. Trajectory forecasting enables sub-routines such as collision avoidance to assess the situation and react accordingly in time. This paper investigates Transformer and Performer models for trajectory prediction of surface vessels in confined waters. Using widely available position data from the Automatic Identification System (AIS), candidate models are trained and evaluated. Historical information on how other vessels have moved in the local area surrounding each agent is conveyed using positional heat maps. The heat maps are generated using Kernel Density Estimation (KDE) and encoded with a Convolutional Neural Network (CNN) before entering the Transformer models. The models are trained with quantile loss to efficiently obtain a model-aware prediction interval, as this is crucial for timely collision avoidance.
The proposed model architectures are shown to greatly outperform previous work while the inclusion of local heat maps enables the prediction of vessel movement that is not otherwise indicated from the previous positions.

\end{abstract}

\input{01_introduction}
\input{02_relatedWork}

\input{03_data}

\input{05_model}

\input{04_training}
\input{06_results}
\input{07_conclusion}

\section*{Acknowledgements}
This work was sponsored by the Danish Innovation Fund, The Danish Maritime Fund, Orients Fund and the Lauritzen Foundation through the ShippingLab project (grant 8090-00063B).

\bibliographystyle{unsrtnat}
\bibliography{references}

\end{document}

%% file: 01_introduction.tex
\section{Introduction}
Recently, autonomous surface vessels have gained notable research interest. Services such as decision support or full autonomy are envisaged to be a crucial part of the future of shipping. This, however, requires excellent situational awareness \cite{Blanke_et_al:2019a,Porathe:2014}.

Situational awareness is obtained through three main steps: perception, understanding of the environment, and anticipation of how the situation develops \cite{endsley:1995}. While nearby vessels can be detected on both radar and cameras \cite{Stets:2019}, estimation of future trajectories is needed if a short term path should be planned. A human navigator will use information about surrounding waters from the Electronic Chart Display and Information System (ECDIS) together with the past motion of vessels to assess the current situation. The past motion of nearby ships can be obtained e.g. by detection and tracking using radar \cite{perera2012maritime,wilthil2017target} or tracking camera detections \cite{Schoeller:2020,Prasad:2017}. Commercial vessels and some large leisure crafts use Automatic Identification System (AIS) to broadcast current position, speed and heading among others.

Trajectory forecasting is a widely researched area, especially for pedestrian motion prediction. Many methods use Long short-term memory (LSTM) networks for motion prediction such as \cite{Kosaraju:2019,Sadeghian:2019,Gupta:2018}. While Recurrent neural networks (RNN) are commonly used for tasks such as time series forecasting and Natural Language Processing (NLP), Transformer networks, first described in \cite{Vaswani:2017}, have recently had immense success, especially for NLP, where models such as Bidirectional Encoder Representations from Transformers (BERT) \cite{BERT}, BART \cite{lewis2019bart} and GPT-3 \cite{brown2020language} outperformed previous LSTM models. Transformer networks have also shown their strengths in pedestrian trajectory forecasting \cite{Giuliari:2020}, where they were shown to outperform the previous LSTM state of the art. One drawback of Transformers is that the computational burden increases quadratically with input sequence length. Multiple recent studies have investigated methods to introduce Transformer models with linear complexity \cite{wang2020linformer,kitaev2020reformer}, though these often do not match the performance of the original Transformer. One study introduced the Performer \cite{Choromanski:2020} that incorporates Fast Attention Via Positive Orthogonal Random Features (FAVOR+) and obtains linear complexity. Their \emph{Performer-ReLU} model showed to outperform a similar Transformer configuration with quadratic complexity while being more efficient.

If a trajectory forecast is to be used for risk assessment in the area of short term path planning for autonomous vessels, it is crucial not to obtain a single prediction, but to produce a prediction interval in which the observed vessel could go. A prediction interval can be obtained in numerous ways. For example by using Monte Carlo dropout \cite{Gal:2016}, using Bayesian neural networks with probabilistic layer weights \cite{mackay1995bayesian} or training the model as a conditional generative adversarial network (GAN) \cite{mirza:2014}. Common for these methods is that a prediction interval is not obtained directly from the network, but by running the network multiple times to approximate a prediction interval. This might result in a misleading prediction interval, as the network is trained to output a prediction close to the mean prediction and the prediction interval is approximated by drawing from a distribution. To circumvent these problems, this work utilizes quantile loss \cite{Koenker:1978} to estimate prediction quantiles during training, which generates a model that is both aware of its prediction interval and can supply this in a single forward pass.

%% file: 02_relatedWork.tex
\section{Related Work}
\label{seq:relatedwork}
A recent assessment of collision avoidance methods for marine vessels mentions different ways to anticipate the trajectory of a given vessel \cite{Huang2020}. Here, the prediction schemes are categorized into physics-based, interaction-aware, and maneuver-based methods. Physics-based methods usually assume that the target vessel maintains a constant course and speed. A disadvantage of such assumptions lies in the lack of information about the curve of the underlying navigational channels. 

Maneuver-based methods use the information contained within the COLREGs or available historical data, such as AIS. Hexeberg et al. \cite{Hexeberg:2017} described a single-point neighbor search method for long-term trajectory prediction using AIS data. Here, data points within a set distance to the target vessel were used to estimate the course and speed for the next timestep if the used points had a similar heading. The method could therefore model the mean trajectory for a vessel. A drawback to such an approach is the difficulty of predicting a trajectory that differs from the usual fairways. Rong et al. \cite{rong:2019} fitted a Gaussian process on AIS data for a southbound shipping route. The Gaussian process was used to predict the lateral movement at discrete time steps while historical acceleration data was used for longitudinal movement estimation. Murray et al. \cite{Murray:2020} used clustered AIS data to predict the future trajectory of a target vessel based on cluster points near that of the target. An autoencoder approach was utilized for the generation of future movement by exploring the latent space, thereby generating a prediction interval. By using unsupervised learning \cite{Pallotta2013} developed a method for AIS data that proposes potential routes and endpoints for the target vessel. The proposed routes are given a probability that is updated iteratively during run time. Tang et al. \cite{tang2019} used a LSTM to forecast the future position of a marine vessel using AIS data. The LSTM was trained to output the parameters of a bivariate gaussian distribution, describing the potential position of a vessel at a chosen point in the future relative to the amount of input data used. Sch\"oller et al. \cite{fets:2021} used an LSTM-GAN to forecast marine vessel trajectories using AIS data. The network had an auxiliary input consisting of CNN features from an AIS heat map of the area surrounding the vessels. It was found that including information about how previous vessels had moved in a given area would increase performance significantly enabling the method to predict movement which would not otherwise be indicated in the previous positions. 

In the domain of pedestrian trajectory prediction, \cite{Gupta:2018} proposed a method for encoding interaction between agents called social pooling by including other agents within a grid. The pooling method was used in training an LSTM-GAN and enabled the generation of socially aware trajectories. Sadeghian et al. \cite{Sadeghian:2019} extended this idea by modeling social interactions using attention. Furthermore, a CNN was included to model the physical features of the environment. The most recent addition to the social GANs, \cite{Kosaraju:2019}, used a graph attention network \cite{velickovic2018graph} for social interaction which had the advantage of being invariant to the number of agents in the scene. Even though the methods with social interaction obtained better results than similar methods without social encoding, \cite{Giuliari:2020} argued that using the original Transformer network (\cite{Vaswani:2017}) for time series forecasting yielded better results, even without social modeling, obtaining state of the art results on common pedestrian trajectory data sets. Giuliari et al. \cite{Giuliari:2020}, obtained prediction intervals by using a quantized version of the Transformer that performed step classification instead of regression, thereby enabling probabilistic modeling from the \emph{softmax} output. 

\todo{The bottlenecks of current AIS-based vessel trajectory prediction schemes are several:
Some current methods only take characteristics of the investigated vessel into account, environmental factors such as water depth, land and navigational fairways are thus unknown and not modeled. Situations can therefore occur where a trajectory is predicted to go through land, which is infeasible. Other methods have the opposite problem. Clustering methods solely rely on the past movements of other vessels in the local area. The predicted trajectory will thus be some average of the trajectories of other similar vessels without directly taking the past movement of the investigated vessel into account.
Finally, existing models use sampling-based approaches to obtain probabilistic predictions, which drastically increase inference time.

This paper, therefore, proposes a hybrid approach using Transformer models that uses the movement characteristics of the individual investigated vessel while considering the positional probability distribution of the local area. 
Most time-series-based approaches to ship trajectory prediction rely on recurrent neural networks such as LSTM \cite{hochreiter:1997} and Gated Recurrent Unit (GRU) \cite{chung2014empirical}. Even though Transformer models have gained immense popularity in recent years, they have, to the best of the author's knowledge, not been investigated in the context of trajectory prediction for marine vessels. 
 To forecast future trajectories, the proposed method uses a vessel's past movements together with a novel approach to including spatial probability distributions in form of positional heat maps in Transformer model. A probabilistic output is supplied without the need for Monte Carlo sampling by using quantile loss. The model therefore directly outputs a prediction interval useful for risk assessment.}

The rest of this paper is structured as follows: Section \ref{seq:relatedwork} discusses the related work done in the area of trajectory forecasting. Section \ref{seq:data} introduces the data used in this research. Section \ref{sec:model} describes the model architectures investigated in this work and explains the difference between the Transformer and the Performer. Section \ref{sec:training} lists the training parameters used and introduces the loss function. Section \ref{sec:results} introduces the performance metrics used for evaluation and discusses the results. Finally, Section \ref{sec:conclusion} presents the conclusions.

%% file: 03_data.tex
\section{Data}
\label{seq:data}
The data processed in this work comes from AIS messages. AIS is an automatic tracking system in which ships send out relevant information about themselves. AIS messages include information such as unique identification (MMSI), position, course over ground, speed, and destination harbor. Vessels of more than 300 gross tonnage, passenger ships, and fishing vessels with a length above 15 meters are obligated to have an AIS transceiver installed, while recreational crafts can choose whether or not to install AIS. The frequency of dynamic messages varies from once every 2 seconds to once every 3 minutes depending on vessel speed and activity. Static information is broadcast once every 6 minutes. The AIS messages used in this work are recorded in Danish waters and are publicly available. 

For the application of trajectory forecasting, the AIS messages are filtered to include only the position of the given vessel in GPS coordinates. While other attributes such as speed over ground, heading, course over ground, length, width, and draft affect the future movements of a vessel, \todo{it was chosen to solely rely on positional data to obtain a more general approach that can be transferred to using other sensors during deployment such as radar and visual tracking. Furthermore, information such as speed and heading is easily derived from the positional data by the model}.

\begin{figure}[htb]
    \centering
    \includegraphics[width=0.55\linewidth]{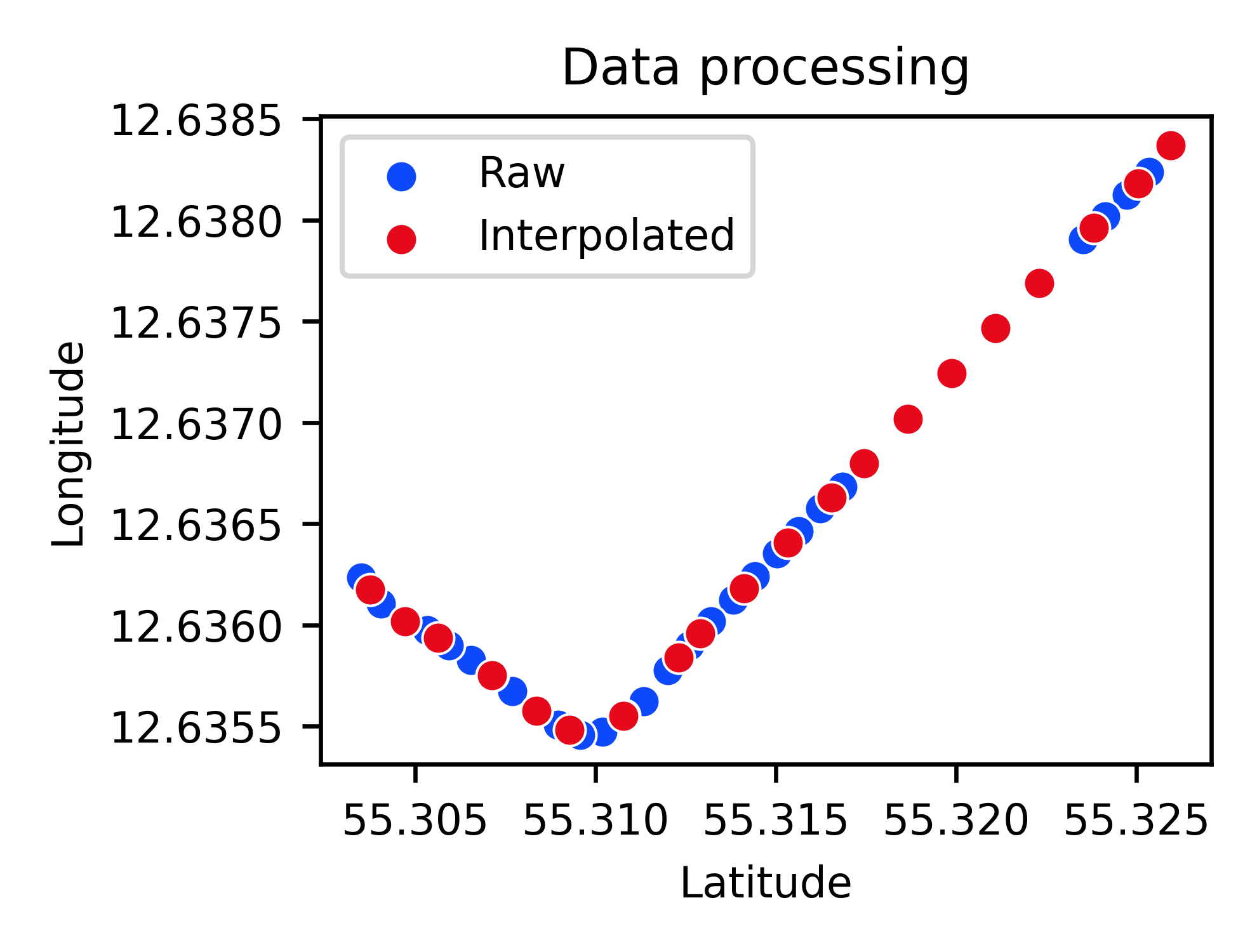}
    \caption{Example of linear interpolation of positional AIS data.}
    \label{fig:interpolation}
\end{figure}

The raw AIS data is processed by first removing tracks with an average velocity below 0.5 knots to reject stationary objects such as buoys. As mentioned, the frequency of AIS messages varies depending on vessel activity. The methods discussed in this paper expect temporally equidistant data, which is why the positional data is resampled using linear interpolation, such that each data point is 20 seconds apart. Using the timestamp of each point, samples are resampled to the nearest 20-second mark, weighted by temporal distance to said mark. Temporal gaps are linearly filled from the nearest resampled data points as seen in Figure~\ref{fig:interpolation}. For deployment, the method suggested in this paper will run only when a new AIS message arrives such that a trajectory is not predicted using extrapolated data. 

When two vessels approach each other, the smallest distance between them during the encounter is called Closest point of approach (CPA). For a given situation, the time left before CPA is called time to CPA (TCPA). Given that a navigator would like to be able to react to a potentially hazardous situation within a TCPA of 10 minutes, the models assessed in this paper will produce an output of 30 predicted timesteps, corresponding to 10 minutes.

Specifically, given a set of $n$ vessels within an area, for a vessel $i$, the $t_{obs}$ most recent positions are provided as ${\mathbf{X}_i = \{x_t^{(i)}\}_{t=0}^{t_{obs}}}$, in Cartesian coordinates $\mathbf{x} \in \mathbb{R}^{2}$. It is desired to forecast the future positions of the vessel ${\hat{\mathbf{Y}}_i=\{x_t^{(i)}\}_{t=t_{obs}}^{t_{obs}+T_{pred}}}$ for $T_{pred}$ predicted time-steps, where $t_{obs} = 15$ and $T_{pred} = 30$. \todo{To make the prediction invariant to global position of the vessels, the latitude and longitude position at each time step in $\{x_t^{(i)}\}_{t=0}^{t_{obs}+T_{pred}}$ is converted to relative coordinates in meters relative to $ \{x_{t_{obs}}^{(i)}\}$. The conversion from GPS coordinates to relative coordinates is done using the haversine formula.}

\begin{figure*}[htb]
    \centering
     \begin{subfigure}[c]{0.28\textwidth}
         \centering
         \includegraphics[width=\textwidth]{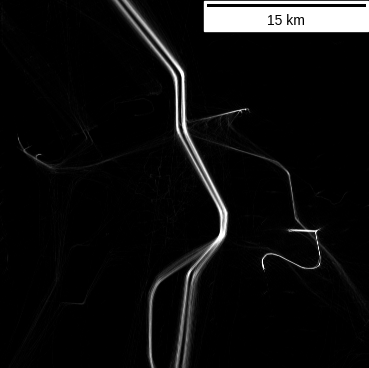}
         \caption{The Great Belt}
         \label{fig:routet}
     \end{subfigure}
     \begin{subfigure}[c]{0.28\textwidth}
         \centering
         \includegraphics[width=\textwidth]{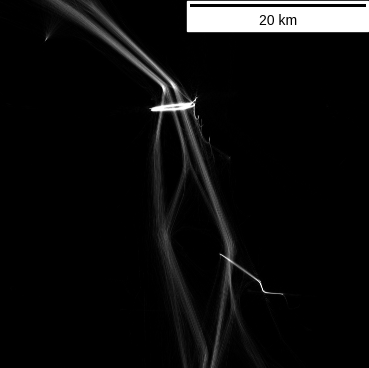}
         \caption{Hven}
         %\label{fig:routes}
     \end{subfigure}
     \begin{subfigure}[c]{0.28\textwidth}
         \centering
         \includegraphics[width=\textwidth]{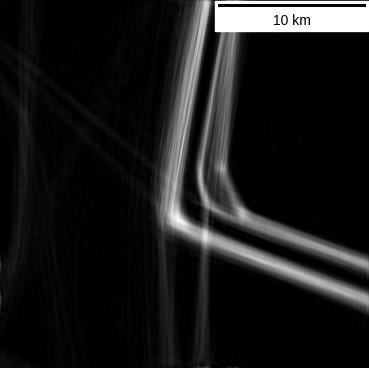}
         \caption{Falsterborev}
         %\label{fig:routes}
     \end{subfigure}
   
     \caption{Heat maps of AIS data from the areas around The Great Belt, Oresund, and Falsterborev. The heat maps are generated using kernel density estimation.}
     \label{fig:kde}
     
\end{figure*}

The AIS messages used in this paper are sampled in the areas near various heavily trafficked trade routes in Danish waters. The areas are
\begin{itemize}
    \item \emph{The Great Belt:} between the two islands of Zealand and Funen.
    \item \emph{Hven:} The area around the Swedish island Hven, where traffic goes between Denmark and Sweden.
    \item \emph{Falsterborev:} The area south of Sweden where traffic from the Baltic Sea passes through.
\end{itemize} 
The areas from which the data is collected have been selected due to a large amount of traffic, and due to the fairways of the areas having sudden turns.
The data is collected over the span of one month. Figure \ref{fig:kde} shows heat maps generated using AIS messages from the areas. Here, the shipping routes and fairways from the sea charts are clearly seen as the white lines. The sudden turns seen in the heat map come from the deep-water routes described in the sea charts. The black space between the fairways is the separation zone of the particular route. The heat map shows that there is a clear tendency of vessels with AIS following the fairways. A model with awareness of how previous vessels have moved in the local area is therefore preferable.

Table \ref{tab:train_data} shows an overview of the training data used in this paper.

\begin{table}[htb]
    \centering
    \small
    \caption{Overview of the training data.}
    \label{tab:train_data}
    \begin{tabular}{llrl}
        \toprule
        \textbf{Area} & \textbf{Bounding rectangle} & \textbf{\# Trajectories} & \textbf{Period} \\
        \midrule
        The Great Belt & \makecell[l]{$55^\circ 4.2259'$N, $10^\circ 0.1282'$E \\ $55^\circ 26.8333'$N, $11^\circ 19.216'$E} & 574\,171 & September 2018 \\
        \addlinespace
        Hven & \makecell[l]{$55^\circ 47.9365'$N, $12^\circ 23.3764'$E \\ $56^\circ 8.0261'$N, $13^\circ 3.6935'$E} & 653\,979 & May 2018 \\
        \addlinespace
        Falsterborev & \makecell[l]{$55^\circ 12.6989'$N, $12^\circ 27.1789'$E \\ $55^\circ 26.144'$N, $12^\circ 50.4963'$E} & 408\,070 & September 2018 \\
        \bottomrule
    \end{tabular}
\end{table}

\begin{comment}
\begin{figure}[H]
    \centering
    \includegraphics[width=0.8\linewidth]{kde_img.png}
    \caption{Heat map of AIS data from the area around The Great Belt. The heat map is generated using kernel density estimation. }
    \label{fig:kde}
\end{figure}

\begin{figure}[H]
    \centering
    \includegraphics[width=0.8\linewidth]{oresund.png}
    \caption{Heat map of AIS data from the area around Oresund.}
    \label{fig:oresund}
\end{figure}
\end{comment}

%% file: 05_model.tex
\section{Transformer type networks}
\label{sec:model}
The Transformer architecture is used to predict the future motion of an agent based on its past movement together with a positional probability distribution from the area. This section describes the inputs and outputs of the model, the general encoder-decoder Transformer structure, the difference between the Transformer and the Performer, and how historical AIS data is included in the models. 

\subsection{Input \& Output}
Transformer models were originally developed for NLP, where each input word is embedded into a floating-point value of a higher dimensional space using a learned lookup table with a known dictionary size. As this paper operates on positional data, a dictionary size is not defined. The inputs are therefore embedded to a \textit{D}-dimensional space using a linear transformation $e_{obs}^{(i,t)}=x_t^{(i)^T} \mathbf{W^x}$, where $\mathbf{W^x}$ is a learnable weight matrix.
The output of the model is the probabilistic prediction quantiles for the future trajectory of the $i$-th agent $\hat{\mathbf{Y}}_{i}=e_{dec}^{(i)}\mathbf{W^{d}}$, where $e_{dec}$ is the output embedding of the model decoder.

\subsection{Model Architecture}
\begin{figure*}[tb]
    \centering
    \includegraphics[width=0.8\textwidth]{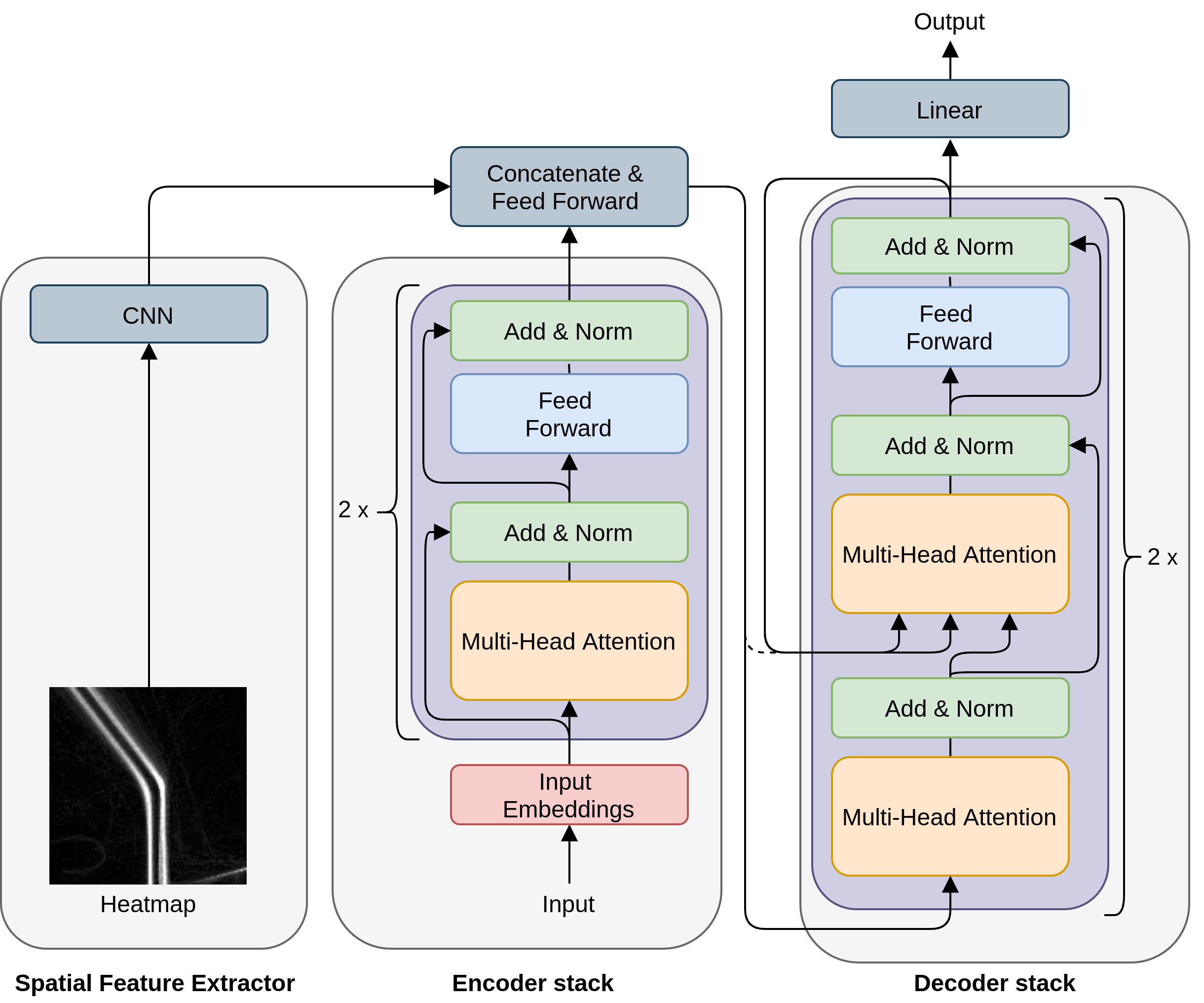}
    \caption{Model illustration of the implemented Transformer architecture.}
    \label{fig:model}
\end{figure*}

 The model has three main components, an encoder, a convolutional neural network (CNN) for spatial feature extraction and a decoder which are described below. An illustration of the model architecture is shown in Figure \ref{fig:model}.

 \paragraph{Encoder.} \todo{The encoder is tasked with encoding the temporal features of a vessel's movement into a vector representation. The input of the encoder stack is the past positions of the observed vessels $e_{obs}^{(i,t)}$ projected into $d_{model}$-dimensional space using a linear projection. The encoder consists of a stack of $N$ layers. Each layer has a Multi-Head Attention step followed by a feed-forward neural network (FFNN) with ReLU activation as suggested by Vaswani et al.~\cite{Vaswani:2017} Each step is followed by layer normalization with a residual connection, denoted by the ``Add \& Norm'' block in Figure \ref{fig:model}. All sub-layers in the model have an output dimension of $d_{model}$. }

 \paragraph{Spatial feature extractor.} 
 It is desired to include information about the marine fairways in the local area around each vessel, as these can indicate an upcoming trajectory change that past movement may not convey. In this work, such information is included in the form of an image which is passed to a CNN. One way of generating an image of the underlying fairways is by looking at the heat map of vessel locations, as large vessels are bound to primarily follow the fairways. Kernel density estimation (KDE) is suitable for generating such an image as it is a non-parametric way to estimate the probability density function of the data, thereby generating a heat map. The heat map is generated by a 2D KDE using the fastKDE algorithm \cite{OBRIEN2016}. A heat map is generated for the entire area of the data and is cropped to the local area around a given vessel, as only the local area influences short term future movement. 
 The cropped heat map of the local area is used as input to a CNN which encodes the spatial features into a vector representation. The output of the spatial feature extractor is concatenated with the encoded temporal features from the encoder stack and projected to $d_{model}$ dimensional space using a linear projection.

  \paragraph{Decoder.} \todo{The decoder takes the merged spatial and temporal information from the spatial feature extractor and encoder and uses it to predict a future trajectory with a $95\%$ prediction interval. 
  The decoder consists of a stack of $N$ layers. The merged features from the encoder stack are passed through two multi-head attention layers followed by a FFNN. The second multi-head attention step works in a recurrent fashion by being fed the output of the previous decoder layer. 
  A linear transformation converts the output of the decoder from model dimension to the three prediction quantiles, each in cartesian coordinates, $\hat{\mathbf{Y}}_{i}=e_{dec}^{(i)} \mathbf{W^{d}}$, where $e_{dec}$ is the output embeddings produced by the decoder stack and $\mathbf{W^{d}}$ is a learnable weight matrix.}

\paragraph{Multi-head attention.} Attention is essentially a map between a query and a set of key and value pairs that produce an output. The output consists of a weighted sum of the values, weighted based on a compatibility function between the query and the corresponding key. \todo{By weighting the inputs, the attention module enables the model to focus on the parts of the input that are most important to the task.} In order to be true to the original Transformer paper \cite{Vaswani:2017}, the attention type used in this particular case is \emph{Scaled Dot-Product Attention} which is defined in Equation \ref{eq:attention}.

\begin{equation}
\operatorname{Attention}(\mathbf{Q}, \mathbf{K}, \mathbf{V})=\operatorname{softmax}\left(\frac{\mathbf{Q} \mathbf{K}^{\top}}{\sqrt{d}}\right)\mathbf{V},
\label{eq:attention}
\end{equation}
where $\mathbf{Q}$ contains the queries, $\mathbf{K}$ the keys, $\mathbf{V}$ the values and $d$ is the dimension of the aforementioned. Here, the weighting is computed using the \emph{softmax} function 
\begin{equation}
 \operatorname{softmax}(z) =\frac{e^{z_{i}}}{\sum_{j=1}^{K} e^{z_{j}}}, \text{for } i=1, \ldots K
 \label{eq:softmax}
\end{equation}
where $K$ is the number of entries in $z$.
Instead of having a single attention layer, the Transformer model uses multi-head attention. Here, the queries, keys and values are first linearly projected $h$ times to $d_k$, $d_k$ and $d_v$ dimensionality respectively. The attention mapping is then run on each of the $h$ heads in parallel and the outputs concatenated, as described in Equation \ref{eq:multihead}.

\begin{equation}
\begin{aligned}
\operatorname{MultiHead}(\mathbf{Q}, \mathbf{K}, \mathbf{V}) &=\text {Concat}\left(\text {head}_{1}, \ldots, \text {head}_{\mathrm{h}}\right) \mathbf{W^{O}} \\
\text {where head}_{\mathrm{i}} =\text{Atten}&\text{tion}\left(\mathbf{Q W_{i}^{Q}}, \mathbf{K W_{i}^{K}}, \mathbf{V W_{i}^{V}}\right)
\end{aligned}
\label{eq:multihead}
\end{equation}

\paragraph{Performer models.} Performer models differ from the traditional Transformer by using \emph{Fast  Attention  Via  positive  Orthogonal  Random  features} (FAVOR+) attention instead of the regular \emph{Scaled Dot-Product Attention}. While the \emph{Scaled Dot-Product Attention} is quadratic in complexity due to the $\mathbf{QK^T}$ expression in equation \ref{eq:attention}, \cite{Choromanski:2020} introduced the linear FAVOR+ attention method, in which the attention matrix is computed by decomposition using kernel functions. Normal decomposition is not possible due to the exponential in the numerator in Equation \ref{eq:softmax}. \emph{Scaled Dot-Product Attention} can be rewritten to the form:
\begin{equation}
\begin{aligned}
\operatorname{Attention}(\mathbf{Q}, \mathbf{K}, \mathbf{V})&=\mathbf{D}^{-1} \mathbf{A} \mathbf{V},\\
\mathbf{A}&=\exp \left(\mathbf{Q} \mathbf{K}^{\top} / \sqrt{d}\right)\\
\mathbf{D}&=\operatorname{diag}\left(\mathbf{A} \mathbf{1}_{L}\right),
\end{aligned}
\label{eq:attention2}
\end{equation}
where $\mathbf{A} \in \mathbb{R}^{L \times L}$ is the \emph{attention matrix} and $\mathbf{1}_{L}$ is the all-ones vector of length $L$. In Equation \ref{eq:attention2}, the \emph{softmax} function is implicitly calculated during the $\mathbf{D}^{-1} \mathbf{A}$ operation. In order to decompose $\mathbf{A}$, FAVOR+ use attention matrices on the form $\mathbf{A}(i, j)=\mathrm{K}\left(\mathbf{q}_{i}^{\top}, \mathbf{k}_{j}^{\top}\right)$, where $\mathbf{q}_{i}$ and $\mathbf{k}_{j}$ is the $i^{t h}$ query and  $j^{t h}$ key vector in $\mathbf{Q}$ and $\mathbf{K}$ respectively. The kernel, K: $\mathbb{R}^{d} \times \mathbb{R}^{d} \rightarrow \mathbb{R}_{+}$ is defined for the randomly sampled mapping $\phi: \mathbb{R}^{d} \rightarrow \mathbb{R}_{+}^{r}$: 
\begin{equation}
\mathrm{K}(\mathbf{x}, \mathbf{y})=\mathbb{E}\left[\phi(\mathbf{x})^{\top} \phi(\mathbf{y})\right]
\end{equation} 
With $\mathbf{Q'}$ and $\mathbf{K'}\in \mathbb{R}^{L \times r}$ having rows $\phi(\mathbf{q}_{i}^{\top})^\top$ and $\phi(\mathbf{k}_{i}^{\top})^\top$ respectively, an approximation of the attention mechanism with linear complexity is given by

\begin{equation}
\begin{aligned}
\widehat{\operatorname{Attention}}(\mathbf{Q}, \mathbf{K}, \mathbf{V})&=\widehat{\mathbf{D}}^{-1}\left(\mathbf{Q}^{\prime}\left(\left(\mathbf{K}^{\prime}\right)^{\top} \mathbf{V}\right)\right), \\ \widehat{\mathbf{D}}&=\operatorname{diag}\left(\mathbf{Q}^{\prime}\left(\left(\mathbf{K}^{\prime}\right)^{\top} \mathbf{1}_{L}\right)\right)
\end{aligned}
\end{equation}

By choosing $\phi$ on the form presented in Equation \ref{eq:phi}, it is possible to model most kernels and not only the before-mentioned \emph{softmax} function. \cite{Choromanski:2020} exploits this by introducing a Transformer model with \emph{ReLU} activation instead of \emph{softmax} by setting $f=ReLU(x)$ in Equation \ref{eq:phi}. They call this the \emph{Performer-ReLU} model and find it to outperform the traditional Transformer presented in \cite{Vaswani:2017}.

\begin{equation}
\begin{aligned}
    \phi(\mathbf{x})=\frac{h(\mathbf{x})}{\sqrt{m}}(&f_{1}\left(\omega_{1}^{\top} \mathbf{x}\right), \ldots, f_{1}\left(\omega_{m}^{\top} \mathbf{x}\right), \ldots, \\
    &f_{l}\left(\omega_{1}^{\top} \mathbf{x}\right), \ldots, f_{l}\left(\omega_{m}^{\top} \mathbf{x}\right)) 
\label{eq:phi}
\end{aligned}
\end{equation}
where $h(x)$ and $f_1 \ldots f_l$ are a set of chosen functions, and $\omega_1, \ldots, \omega_m$ are random feature vectors sampled independently from a distribution $\mathcal{D}$ (typically isotropic Gaussian, optionally orthogonalized).

%% file: 04_training.tex
\section{Training Specifications}
\label{sec:training}
The models were trained on 1\,636\,220 trajectories, collected over the span of a month. The trajectories were collected from the different areas presented in Table \ref{tab:train_data}.
The AdamW optimizer \cite{loshchilov2019adamw} was used together with a warm restart learning rate scheduler \cite{loshchilov2017sgdr} with a base learning rate of $\alpha = 1\mathrm{e}{-4}$, weight decay of $1\mathrm{e}{-2}$ and $T_0 = 0$, $T_{mult}=2$ as parameters for the scheduler as recommended in \cite{loshchilov2019adamw} and \cite{loshchilov2017sgdr} respectively.

\todo{Both the Transformer and the Performer models have internal dimension $d_{model} = 128$, $h=8$ attentional heads with query and value of dimension $d_v, d_q = 16$. }
Internal dimension was chosen conservatively as the problem at hand is computationally simple compared to the translation task that the original Transformer was built for. $h$, $d_v$ and $d_q$ are chosen with respect to $d_{model}$ as described in \cite{Vaswani:2017}. The spatial feature extractor is a small CNN, with the structure presented in Table \ref{tab:CNN}. \todo{The input size of $64\times64$ was found to be sufficient to convey the positional probability density of the local area. A local heat map is generated by cropping the larger heat maps shown in Figure \ref{fig:kde}. Given the past positions of a vessel, the crop is carried out such that the latest vessel position is centered in the local heat map. The local heat map covers an area of $8000\times8000$ meters.}

\begin{table}[htb]
    \centering
    \small
    \caption{Structure of the spatial feature extractor. Each \emph{Conv2d} block is followed by batch normalization and \emph{LeakyReLU} activation.}
    \label{tab:CNN}
    \begin{tabular}{lccccc}
        \toprule
        \textbf{Operation} & \textbf{\# Filters} & \textbf{Kernel} & \textbf{Stride} & \textbf{Padding} & \textbf{Output size} \\
        \midrule
        Input & -- & -- & -- & -- & $64\times64\times1$ \\
        \midrule
        Conv2d & 8 & 3 & 1 & 1 & $64\times64\times8$ \\
        Conv2d & 8 & 3 & 1 & 1 & $64\times64\times8$ \\
        Conv2d & 8 & 3 & 2 & 1 & $32\times32\times8$ \\
        \midrule
        Conv2d & 16 & 3 & 1 & 1 & $32\times32\times16$ \\
        Conv2d & 16 & 3 & 1 & 1 & $32\times32\times16$ \\
        Conv2d & 16 & 3 & 2 & 1 & $16\times16\times16$ \\
        \midrule
        Conv2d & 32 & 3 & 1 & 1 & $16\times16\times32$ \\
        Conv2d & 32 & 3 & 1 & 1 & $16\times16\times32$ \\
        Conv2d & 32 & 3 & 2 & 1 & $8\times8\times32$ \\
        \midrule
        Flatten & -- & -- & -- & -- & $2048$ \\
        Linear & -- & -- & -- & -- & $256$ \\
        \bottomrule
    \end{tabular}
\end{table}

As trajectory forecasting is meant to be used for decision support or in an autonomous system, it is desired to obtain a probabilistic prediction interval to gain insight into the uncertainty of the model and for risk assessment. In this paper, such an interval is obtained by training the models with \emph{quantile loss} as described in Equation \ref{eq:loss}.

\begin{equation}
    L(\mathbf{\hat{Y}}_i, \mathbf{Y}_i) = \frac{1}{Q} \sum_{j=1}^Q{\max\left(q_j (\mathbf{Y}_i - \mathbf{\hat{Y}}_{i,j}), (q_j-1)(\mathbf{Y}_i - \mathbf{\hat{Y}}_{i,j})\right)}
    \label{eq:loss}
\end{equation}
where $Q$ is the number of quantiles in $q=[0.025, 0.5, 0.975]$. Using quantile loss, the models will have three outputs, one for each quantile. Each of the three outputs will consist of the predicted trajectory for the given quantile. The training data is split such that $10\%$ is set aside and used for validation. Training is stopped when the validation loss has converged.

\begin{comment}
The spatial feature extractor of the model architecture is pre-trained as a convolutional autoencoder (CAE) on heavily augmented heat map images. The CAE architecture consists of an encoder and a decoder. The encoder learns to encode the heat maps into compact vector representation while the decoder learns to reconstruct the images from the encoded representation. After pre-training, the decoder part of the CAE is discarded and only the encoder is used during trajectory prediction. During the training of the Transformer model, the weights of the spatial feature extractor are frozen, meaning they will not be updated. This is done to prevent overfitting to the heat maps of the training area and improve generalization.  
\end{comment}

%% file: 06_results.tex
\section{Results}
\label{sec:results}
\begin{figure}[htb]
    \centering
    \begin{minipage}[c]{0.64\textwidth}
        \centering
        \small
        \setlength\tabcolsep{4pt}
        \captionof{table}{Overview of the test data.}
        \label{tab:test_data}
        \begin{tabular}{llrl}
            \toprule
            \textbf{Area} & \textbf{Bounding rectangle} & \textbf{Count} & \textbf{Period} \\
            \midrule
            The Great Belt & \makecell[l]{$55^\circ 4.2259'$N, $10^\circ 0.1282'$E \\ $55^\circ 26.8333'$N, $11^\circ 19.216'$E} & 20\,000 & May 2018 \\
            \addlinespace
            Falsterborev & \makecell[l]{$55^\circ 12.699'$N, $12^\circ 27.1789'$E \\ $55^\circ 26.144'$N, $12^\circ 50.4963'$E} & 20\,000 & May 2018 \\
            \addlinespace
            Oresund & \makecell[l]{$55^\circ 30.6254'$N, $12^\circ 33.576'$E \\ $55^\circ 45.1057'$N, $13^\circ 0.3853'$E} & 20\,000 & May 2018 \\
            \bottomrule
        \end{tabular}
    \end{minipage}\hfill
    \begin{minipage}[c]{0.33\textwidth}
        \centering
        \includegraphics[width=0.85\linewidth]{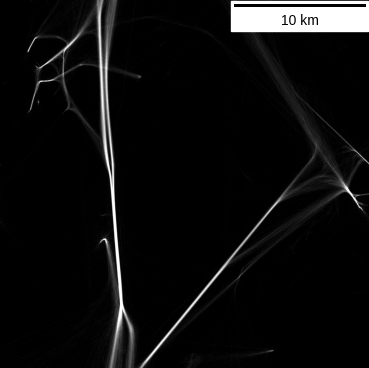}
        \captionof{figure}{Heat map of the Oresund test area, which is not part of the training data.}
        \label{fig:routes}
    \end{minipage}
\end{figure}

This section discusses the performance of the different methods for trajectory prediction. 
Currently, the \emph{automatic radar plotting aid} (ARPA) installed in most large vessels use the assumption that other vessels keep current speed and heading. Two baselines are therefore used for comparison against the proposed methods: One is a Gaussian process, the other is linear regression using the three most recent positions. The linear regression was fitted using quantile regression. Both were chosen as they represent what is currently implemented on large vessels while bringing probability to the predictions. Another comparison was done against the model architecture proposed in \cite{fets:2021}, where an LSTM-GAN was used to forecast the trajectories of marine vessels. Similarly to this paper, the LSTM-GAN used local AIS information in the form of a KDE-generated heat map to aid in the predictions. \todo{Finally, comparison is also done using Transformer models without access to the local heat maps. These models are denoted as $\text{Transformer}_{\text{No Heat}}$ and $\text{Performer}_{\text{No Heat}}$.}
All tested methods have been trained on the data presented in Section~\ref{seq:data}. Evaluation is carried out on three test sets which are presented in Table \ref{tab:test_data}. Two sets are from The Great Belt and Falsterborev areas, though collected during a different time period than in the training set, namely in May 2018. The other test set is gathered from an area not included in the training data. This area, Oresund, is the area around The Oresund Bridge, connecting Denmark and Sweden. The heat map of the Oresund area can be seen in Figure \ref{fig:routes}. For testing, the heat maps used for forecasting are generated using data collected during a different time period than that of the test data as to emulate how the methods would perform in a real-world scenario.

Each method is evaluated using three metrics: average displacement error (ADE), final displacement error (FDE) and interval accuracy (IAcc).

ADE is the average $L^2$ distance in kilometers between the ground truth $Y_i$ and the mean predicted trajectory, $\hat{Y}_{i,2}$, across all time steps.
\begin{equation}
\centering
    \mathrm{ADE}=\frac{\sum_{i=1}^{N} \sum_{t=1}^{T_{pred}}\left\|\hat{Y}_{i,2}^{t}-Y_i^t\right\|_{2}}{N T_{pred}}
    \label{eq:ade}
\end{equation}
where $N$ is the total number of vessels, $T_{pred}$ is the number of timesteps predicted and $\hat{Y}_{i,2}^{t}$ is the predicted position of vessel $i$ for quantile $2$ at timestep $t$.
FDE is the $L^2$ distance in kilometers between $Y_i$ and $\hat{Y}_{i,2}$ at the final time step, given by
\begin{equation}
    \mathrm{FDE}=\frac{\sum_{i=1}^{N}\left\|\hat{Y}_{i,2}^{T_{pred}}-Y_i^{T_{pred}}\right\|_{2}}{N}
    \label{eq:fde}
\end{equation}
where similarly, $N$ is the total number of observed vessels.
Finally, IAcc indicates the fraction of ground truth steps that lie within the quantiles of the predicted trajectory. 
\begin{equation}
    \mathrm{IAcc}=\frac{\sum_{i=1}^{N} \sum_{t=1}^{T_{pred}} acc_i^t}{N T_{pred}}
\end{equation}
where $acc_i^t$ is defined as
\begin{equation}
    acc_i^t = \begin{cases}
        1 & \text{if } \hat{Y}_{i,1}^t < Y_i^t < \hat{Y}_{i,3}^t \\
    0 & \text{otherwise.}	
    \end{cases}
\label{eq:iacc}
\end{equation}
where $\hat{Y}_{i,j}^t$ indicates the predicted position for vessel $i$ at time $t$ for prediction quantile $j$.

\begin{table}[htb]
    \centering
    \small
    \caption{Evaluation results on the three test areas. ADE and FDE are given in kilometres. Inference time is the average time per prediction measured over 1000 batches with a batch size of 1 on an \emph{NVIDIA RTX 3090} GPU. The best result in each group is shown in bold.}
    \label{tab:results}
    \begin{tabular}{lcccc}
        \toprule
        \textbf{Method} & \textbf{ADE} $\downarrow$ & \textbf{FDE} $\downarrow$ & \textbf{IAcc} $\uparrow$ & \textbf{Inference time (ms)} \\
        \midrule
        \multicolumn{5}{l}{\textit{The Great Belt}} \\
        \quad Linear & 0.415 & 1.054 & 7.0\% & 14.5 \\
        \quad GP & 0.534 & 1.303 & 59.5\% & 14.5 \\
        \quad LSTM-GAN & 0.363 & 0.767 & 78.0\% & 157 \\
        \quad Transformer & \textbf{0.137} & \textbf{0.289} & 94.6\% & 6.0 \\
        \quad Performer & \textbf{0.137} & 0.291 & \textbf{94.7\%} & 7.1 \\
        \quad $\text{Transformer}_{\text{No Heat}}$ & 0.392 & 1.024 & 84.0\% & \textbf{4.9} \\
        \quad $\text{Performer}_{\text{No Heat}}$ & 0.416 & 1.069 & 79.9\% & 6.0 \\
        \midrule
        \multicolumn{5}{l}{\textit{Falsterborev}} \\
        \quad Linear & 0.282 & 0.691 & 4.0\% &  \\
        \quad GP & 0.326 & 0.808 & 66.6\% &  \\
        \quad LSTM-GAN & 0.313 & 0.642 & 76.6\% &  \\
        \quad Transformer & 0.119 & 0.257 & 95.2\% &  \\
        \quad Performer & \textbf{0.117} & \textbf{0.252} & \textbf{95.7\%} &  \\
        \quad $\text{Transformer}_{\text{No Heat}}$ & 0.239 & 0.637 & 92.2\% &  \\
        \quad $\text{Performer}_{\text{No Heat}}$ & 0.251 & 0.650 & 89.5\% &  \\
        \midrule
        \multicolumn{5}{l}{\textit{Oresund (unseen area)}} \\
        \quad Linear & 0.279 & 0.657 & 3.0\% &  \\
        \quad GP & 0.322 & 0.736 & 62.0\% &  \\
        \quad LSTM-GAN & 0.458 & 0.944 & 76.7\% &  \\
        \quad Transformer & 0.291 & 0.724 & 82.3\% &  \\
        \quad Performer & 0.302 & 0.755 & 83.1\% &  \\
        \quad $\text{Transformer}_{\text{No Heat}}$ & \textbf{0.264} & \textbf{0.648} & \textbf{88.4\%} &  \\
        \quad $\text{Performer}_{\text{No Heat}}$ & 0.273 & 0.664 & 83.7\% &  \\
        \midrule
        \multicolumn{5}{l}{\textit{Average}} \\
        \quad Linear & 0.325 & 0.801 & 4.7\% &  \\
        \quad GP & 0.394 & 0.949 & 62.7\% &  \\
        \quad LSTM-GAN & 0.378 & 0.784 & 77.1\% &  \\
        \quad Transformer & \textbf{0.182} & \textbf{0.423} & 90.7\% &  \\
        \quad Performer & 0.185 & 0.432 & \textbf{91.2\%} &  \\
        \quad $\text{Transformer}_{\text{No Heat}}$ & 0.298 & 0.770 & 88.2\% &  \\
        \quad $\text{Performer}_{\text{No Heat}}$ & 0.313 & 0.794 & 84.4\% &  \\
        \bottomrule
    \end{tabular}
\end{table}

\todo{Table \ref{tab:results} shows the evaluation results. Both the Transformer and the Performer models achieve an average ADE of around 0.18 km, which is far better than any of the baseline methods. The average final displacement error lies at around 0.4 km, which is around half of that of the baselines. The proposed method also has superior IAcc of $\sim 91\%$ which indicates that the ground truth trajectory falls within the prediction interval in $\sim 91\%$ of the predictions.}

The inference speed of the Transformer type models is also significantly faster than the baselines. The LSTM-GAN, for example, has to run multiple forward passes, in order to approximate the model prediction interval. Though, it does seem like the Performer model is slightly slower than the Transformer even though the attention module of the Performer is linear in complexity compared to the quadratic complexity of the Transformer. The slow-down of the Performer might be a result of the relatively small input length of $15$ steps, at which the computational requirement of the kernel estimation in the FAVOR+ attention exceeds that of the \emph{Scaled Dot-Product Attention}. 

From Table \ref{tab:results} it is clearly seen that including heat map data significantly improves performance in The Great Belt area and Falsterborev areas, where the two Transformer models outperform all other methods. Though it is also noted that in the case of the Oresund area, the Transformer models without access to heat map data perform slightly better than their heat map counterparts. This could indicate overfitting of the spatial feature extraction subnetwork to the training areas, as while the training set includes a vast amount of trajectories, cropping each KDE leads to few unique heatmap images for the spatial feature extractor. To test this hypothesis, the Transformer networks are trained on data from the Oresund area collected during September 2018 and tested on data collected during May 2018. The results are presented in Table \ref{tab:results1}.
\begin{table}[htb]
    \centering
    \small
    \caption{Evaluation results on the Oresund area using models trained on data from the same area gathered during a different month. ADE and FDE are given in kilometres.}
    \label{tab:results1}
    \begin{tabular}{lccc}
        \toprule
        \textbf{Method} & \textbf{ADE} $\downarrow$ & \textbf{FDE} $\downarrow$ & \textbf{IAcc} $\uparrow$ \\
        \midrule
        Transformer & 0.222 & 0.483 & 71.1\% \\
        Performer & \textbf{0.206} & \textbf{0.461} & 75.0\% \\
        $\text{Transformer}_{\text{No Heat}}$ & 0.247 & 0.585 & \textbf{86.6\%} \\
        $\text{Performer}_{\text{No Heat}}$ & 0.257 & 0.608 & 85.3\% \\
        \bottomrule
    \end{tabular}
\end{table}
Table \ref{tab:results1} shows the performance of the Transformer models on the Oresund area if trained on data from the same area collected during a different time period. It is noted that in this case, including local heat map information leads to considerably lower displacement errors, although at the cost of a lower interval accuracy. Based on this result, if the proposed method was to be implemented in an onboard system, it would preferably be pretrained on data from the planned route. This would not be unrealistic as for example ferries have a predetermined route that is seldom changed.

\begin{figure*}[tb]
    \centering
         \begin{subfigure}[t]{0.28\textwidth}
         \centering
         \includegraphics[width=\textwidth]{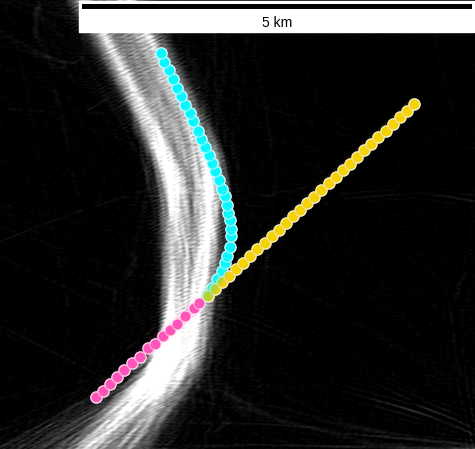}
         \caption{Linear Model}
         \label{fig:lm}
     \end{subfigure}
     \hfill
     \begin{subfigure}[t]{0.28\textwidth}
         \centering
         \includegraphics[width=\textwidth]{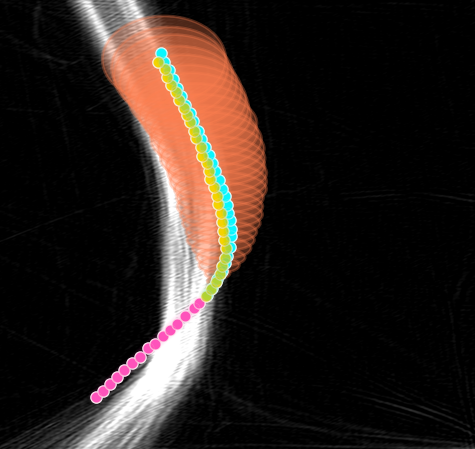}
         \caption{Transformer}
         \label{fig:tkde}
     \end{subfigure}
     \hfill
    \begin{subfigure}[t]{0.28\textwidth}
         \centering
         \includegraphics[width=\textwidth]{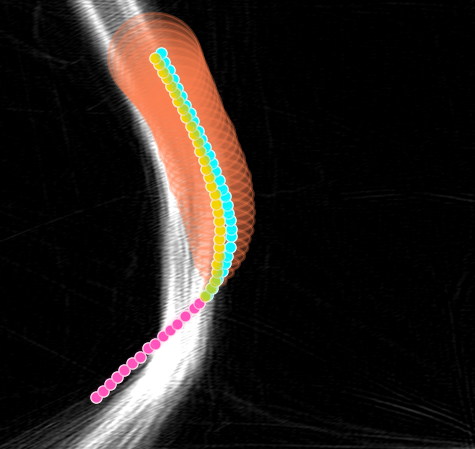}
         \caption{Performer}
         \label{fig:pkde}
     \end{subfigure}
     \hfill
     \begin{subfigure}[t]{0.28\textwidth}
         \centering
         \includegraphics[width=\textwidth]{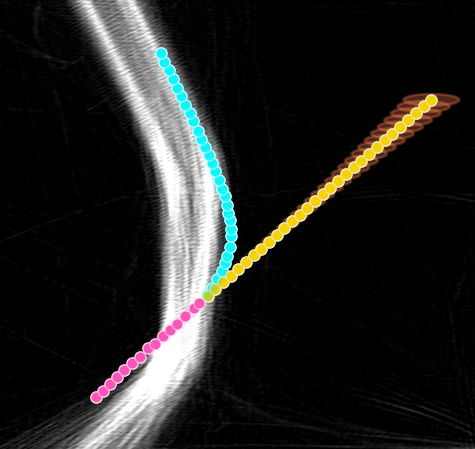}
         \caption{Gaussian Process}
         \label{fig:gp}
     \end{subfigure}
      \hfill
     \begin{subfigure}[t]{0.28\textwidth}
         \centering
         \includegraphics[width=\textwidth]{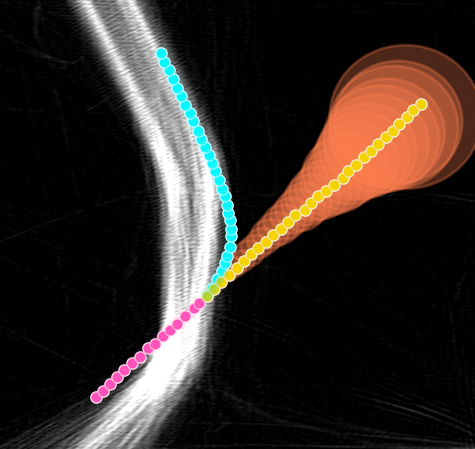}
         \caption{$\text{Transformer}_{\text{No Heat}}$}
         \label{fig:t}
     \end{subfigure}
     \hfill
          \begin{subfigure}[t]{0.28\textwidth}
         \centering
         \includegraphics[width=\textwidth]{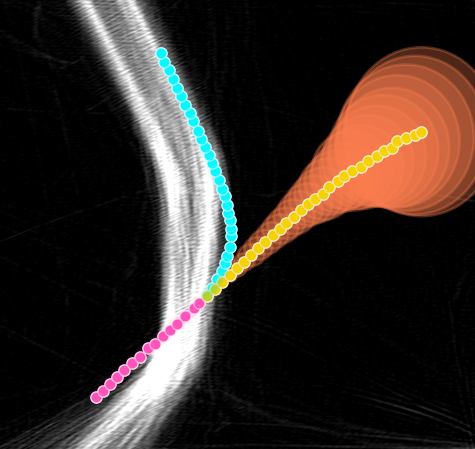}
         \caption{$\text{Performer}_{\text{No Heat}}$}
         \label{fig:p}
     \end{subfigure}
     \begin{subfigure}[t]{0.28\textwidth}
         \centering
         \includegraphics[width=\textwidth]{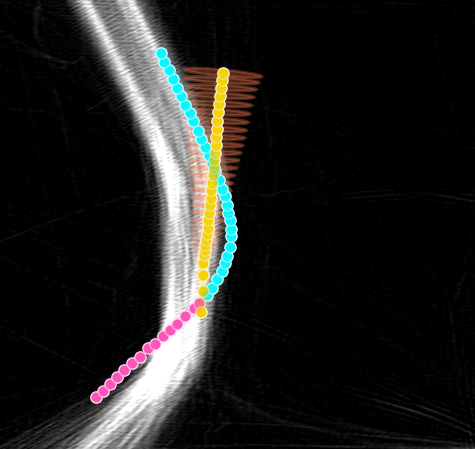}
         \caption{LSTM-GAN}
         \label{fig:lstm}
     \end{subfigure}

     \caption{Trajectory forecasting of a vessel using the different methods. Here, the background shows the local heat map, the pink coordinates indicate the input data, the blue coordinates the true trajectory, the yellow coordinates show the predicted trajectory, and the orange ellipsoids indicate the $95\%$ prediction interval for each time step. It is clearly seen that including the heat map yields a far better prediction. The scale of the sub-figures is shown by the scale bar at the top of Figure~\ref{fig:lm}.}
     \label{fig:results}
\end{figure*}

\todo{To visually show the superiority of the proposed method, forecasted trajectories are plotted against each other in Figure~\ref{fig:results}. The input trajectory is sampled from The Great Belt test set.}

Figure \ref{fig:results} shows the trajectory predictions by the different methods. The trajectories are plotted on top of the local heat map generated using data collected during a different time period. The pink coordinates are the input to the models while the blue coordinates are the ground truth trajectory of the vessel. The yellow coordinates indicate the anticipated trajectory with the $95\%$ prediction interval for each time step being the orange ellipsoids. \todo{The ellipsoids are generated as follows: let $p_{ij}$ describe the point at the i-th latitudinal quantile and j-th longitudinal quantile. The sides of the ellipsoid describing the $95\%$ confidence interval have the lengths $|p_{11}-p_{13}|$ and $|p_{11}-p_{31}|$. The resulting ellipsoid is rotated by the angle between the line from $p_{11}$ to $p_{13}$ and the x-axis. From Figures \ref{fig:tkde} and \ref{fig:pkde} it is clearly evident that the proposed method yields superior predictions. The inclusion of local heat maps enables predictions to follow navigational fairways, correctly predicting the sudden turn. It is furthermore noted that the ground truth trajectory lies well within the prediction interval of the proposed method with little distance to the mean prediction. The LSTM-GAN seems to be trying to model the upcoming turn, but with little success. This might be a result of LSTMs processing data sequentially compared to the parallel processing of the Transformer. The methods without a spatial feature extractor are unable to anticipate the upcoming turn and thus wrongly predict the vessel to keep going straight. Interestingly, the Transformer models without heat map inputs have quite a wide prediction interval for the later steps, which might point to the data set having many sudden turns which cannot be modelled by past positions alone.}

\begin{comment}
Oresund trained on oresund
T_kde:
Acc:  0.709
ADE:  0.219
FDE:  0.481
Loss:  0.142

T:
Acc:  0.518
ADE:  0.457
FDE:  1.114
Loss:  0.316

P_kde:
Acc:  0.741
ADE:  0.207
FDE:  0.466
Loss:  0.127

P
Acc:  0.722
ADE:  0.366
FDE:  0.905
Loss:  0.221

#####################
Oresund

P:
Acc:  0.87
ADE:  0.279
FDE:  0.682
Loss:  0.151

P_KDE
Acc:  0.859
ADE:  0.3
FDE:  0.735
Loss:  0.159

T_KDE
Acc:  0.823
ADE:  0.292
FDE:  0.721
Loss:  0.154

T
Acc:  0.884
ADE:  0.264
FDE:  0.648
Loss:  0.143

RuteT
T:
Acc:  0.884
ADE:  0.264
FDE:  0.648
Loss:  0.143

T_KDE
Acc:  0.944
ADE:  0.141
FDE:  0.297
Loss:  0.066

P:

P_KDE
Acc:  0.952
ADE:  0.142
FDE:  0.303
Loss:  0.066
\end{comment}

%% file: 07_conclusion.tex
\section{Conclusion}
\label{sec:conclusion}

This paper investigated Transformer and Performer models trained with quantile loss for probabilistic trajectory forecasting of marine vessels. A network architecture was proposed that adds a spatial feature extractor to the Transformer, giving the model information on how vessels have previously moved in the local area through KDE-generated heat maps. Training with quantile loss provides a prediction interval in a single forward pass, which makes inference considerably faster than sampling-based GAN-type models (6--7\,ms compared to 157\,ms per prediction).

The Transformer and the Performer performed similarly. Averaged over the three test areas, the Transformer obtained a slightly lower displacement error (ADE of 0.182\,km compared to 0.185\,km), while the Performer obtained a slightly higher interval accuracy (91.2\% compared to 90.7\%). Both models significantly outperformed the previously proposed LSTM-GAN, which had an average ADE of 0.378\,km, as well as the linear and Gaussian process baselines. Despite its linear-complexity attention, the Performer was not faster than the Transformer for the short input sequences of 15 time steps used in this work.

The heat maps were most beneficial in the areas covered by the training data. In the unseen Oresund area, the models without heat maps performed slightly better, indicating that the spatial feature extractor overfits to the training areas. When trained on data from the Oresund area, including heat maps again gave the lowest displacement errors, although at a lower interval accuracy. In practice, the method should therefore preferably be trained on data from the area or route in which it is deployed.

Further work includes improving the generalization of the spatial feature extractor to unseen areas, adding sea chart information to the model, and using positional data from smaller vessels without AIS, such as sailboats.